\documentclass{article}
\usepackage{spconf,amsmath,graphicx}

\usepackage{enumitem}
\setlist{nosep, leftmargin=14pt}

\usepackage{subcaption}
\usepackage{float}
\usepackage{amssymb}
\usepackage{siunitx}
\usepackage{microtype}
\usepackage{xurl}

\graphicspath{{Figures/}}

\DeclareUnicodeCharacter{03B2}{$\beta$}   
\DeclareUnicodeCharacter{2019}{'}         
\DeclareUnicodeCharacter{2212}{-}         

\makeatletter
\@ifpackageloaded{hyperref}{%
  \pdfstringdefDisableCommands{%
    \def\textsuperscript#1{#1}%
    \def\Delta{Delta}%
    \def\beta{beta}%
    \def\alpha{alpha}%
    \def\gamma{gamma}%
  }%
}{}
\makeatother

\title{Forecasting future anatomies: Longitudinal brain MRI-to-MRI Prediction}

\name{%
\parbox{\textwidth}{\centering
Ali Farki, Elaheh Moradi, Deepika Koundal, Jussi Tohka}%
}
\address{A.I. Virtanen Institute for Molecular Sciences, University of Eastern Finland, Kuopio, Finland}

\begin{document}
\maketitle

\begin{abstract}
Predicting future brain state from a baseline magnetic resonance image (MRI) is a central challenge in neuroimaging, and has important implications for studying neurodegenerative diseases such as Alzheimer’s disease (AD). Most existing approaches predict future cognitive scores or clinical outcomes, such as conversion from mild cognitive impairment to dementia. Instead, here  we investigate  longitudinal MRI image-to-image prediction that forecasts a participant's entire brain MRI several years into the future, intrinsically modeling complex, spatially distributed neurodegenerative patterns. We implement and evaluate five deep learning architectures (UNet, U\textsuperscript{2}-Net, UNETR, Time-Embedding UNet, and ODE-UNet) on two longitudinal cohorts (ADNI and AIBL). Predicted follow-up MRIs are directly compared with the actual follow-up scans using metrics that capture global similarity and local differences. The best performing models achieve high-fidelity predictions, and all models generalize well to an independent external dataset, demonstrating robust cross-cohort performance. Our results indicate that deep learning can reliably predict participant-specific brain MRI at the voxel level, offering new opportunities for individualized prognosis.
\end{abstract}


\section{Introduction}
\label{sec:intro}
Understanding how the human brain changes over time is a central goal in neuroimaging research. Longitudinal magnetic resonance imaging (MRI) provides rich insights into brain development, aging, and structural alterations associated with neurological and psychiatric conditions.  In addition, predicting future MRI patterns is valuable because anticipated brain changes can serve as early indicators of disease progression or cognitive decline, offering a data-driven pathway toward forecasting future clinical outcomes. However, longitudinal datasets are often limited in temporal coverage and sample size, constraining our ability to model individual trajectories of brain change. To address these challenges, recent research has begun exploring computational approaches that can generate future brain MRI scans, effectively trying to predict how an individual’s brain structure will evolve over time \cite{xia2021learning,ravi2022degenerative}. Accurate prediction of future brain morphology could open new avenues for studying brain aging and disease progression, improving patient-level prognosis. 

Traditionally, machine learning in neuroimaging has focused on predicting diagnostic \cite{spasov2019parameter} or cognitive outcomes from imaging features \cite{moradi2025integratingnba, colliot2023machine}. Although valuable for clinical applications, these approaches do not offer predictions on how brain anatomy itself changes over time. In contrast, image-to-image prediction seeks to model the full spatial evolution of brain structure, learning how an individual’s MRI transforms over time. This task requires capturing both spatial anatomy and temporal dynamics, making it a challenging and highly informative problem in computational neuroimaging.

A rapidly expanding body of research has explored deep learning approaches for medical image-to-image translation. In neuroimaging, this work has focused primarily on cross-modality synthesis, for example, generating a pseudo-FDG positron emission tomography (PET) image from an MRI or vice versa \cite{dayarathna2024deep,sikka2025mri}. Many MRI synthesis methods using deep learning translate between MRI pulse sequences \cite{dar2019image}, most commonly T1w$\rightarrow$T2w/FLAIR, showing that one sequence can often predict another’s contrast. However, they do not address longitudinal prediction within the same modality. Another related line of research involves MRI-to-MRI image synthesis, in which generative models simulate brain scans under different conditions, such as aging or disease effects, but do not predict the actual future MRI of each individual. Ravi et al. introduced degenerative adversarial neuroimage nets to produce aged brain images constrained by disease patterns \cite{ravi2022degenerative}, and Xia et al. (2021) proposed an autoencoder-based method to generate aged MRIs from a single baseline scan \cite{xia2021learning}. Although these models generate visually convincing progressions, they typically capture average or prototypical changes rather than individualized, subject-specific evolution. More recently, Pombo et al. (2023) combined conditional generative models with diffeomorphic transformations to create “counterfactual” MRIs under altered conditions \cite{pombo2023equitable,fu2025synthesizing}; however, such approaches are not trained to reproduce each participant’s true future brain image.

In this study, we investigate longitudinal MRI-to-MRI prediction, enabling voxel-level modeling of individual brain changes over time. Unlike prior work focused on regional measures \cite{hadji2025predicting} or population-average trajectories\cite{xia2021learning}, our approach predicts the entire future MRI of each individual, capturing the complex spatial pattern of anatomical evolution. In particular, we implement and compare five deep learning architectures within a unified framework, evaluate their predictions against follow-up scans using both global and regional metrics, and test their generalizability on an independent cohort.

\section{Methods}
\label{sec:methods}

\subsection{Datasets and Preprocessing}
Data used in this work were obtained from the ADNI (\url{http://adni.loni.usc.edu}) and the AIBL study group (\url{https://aibl.org.au}). For up-to-date information, see (\url{www.adni-info.org}).  AIBL study methodology has been reported previously \cite{ellis2009australian}. 

We used longitudinal T1-weighted MRI from ADNI and AIBL, including data from all the participants who had undergone at least two scans.  Images were preprocessed using CAT12 \cite{gaser2024cat} with segmentation, spatial normalization, modulation, smoothing (\SI{4}{\milli\metre} FWHM), and downsampling to \SI{4}{\milli\metre\cubed}, yielding gray matter density maps. Two ADNI training sets were generated: BigDataset (2746 pairs from 1255 participants with 24-month intervals at varying starting timepoints) and SmallDataset (1191 baseline-to-month-24 pairs). Data were split by participant (40 validation, 140 test) to prevent leakage. The ADNI test set consisted of baseline-to month-24-pairs.  AIBL served as external validation with 18-month scans extrapolated to 24 months (173 participants).

\subsection{Deep Learning Architectures, Training and Evaluation}
Let $x^i_{t}$ be 3D MRI of participant $i$ at time $t$ expressed in months from baseline. The general problem we solve is as follows: Given a training set $\{(x^i_{t_{k_1}},x^i_{t_{k_2}}): i \in Train, t_{k_1} < t_{k_2}\}$, learn a model $g$ that predicts $x^j_{t_2}$ based on $x^j_{t_1}$, $t_1 < t_2$. In practice, we set a constraint that $t_{k_2} = t_{k_1} + 24, t_2 = t_1 + 24$.   

We implemented five architectures predicting 3D MRI at $t_{k+24}$ from $t_k$: \textbf{UNet} \cite{ronneberger2015u} with encoder-decoder and skip connections; \textbf{U$^2$-Net} \cite{qin2020u2} with nested residual U-blocks for multi-scale features; \textbf{UNETR} \cite{hatamizadeh2022unetr} using Vision Transformer encoder with convolutional decoder; \textbf{Time-Embedding UNet} (TEUNet)\cite{liu2025imageflownet} with learnable time embeddings; \textbf{ODE\mbox{-}UNet}~\cite{chen2018neural,bilovs2021neural} models continuous latent dynamics via $\dot{Z}=f_{\theta}(Z,t)$; integrating this neural ODE from baseline to $t{+}24$ months yields the predicted future scan. Here $Z(t)$ is the latent state, $t$ is time (months), and $f_{\theta}$ is the learned vector field.

UNet, U$^2$-Net, UNETR, and TEUNet used Mean Squared Error (MSE) loss. ODE-UNet employed $\mathcal{L}_{\text{total}} = \mathcal{L}_{\text{MSE}} + \lambda_{\text{feat}} \mathcal{L}_{\text{feat}} + \lambda_{\text{ODE}} \mathcal{L}_{\text{ODE}}$ following ImageFlowNet \cite{liu2025imageflownet}, where $\mathcal{L}_{\text{feat}}$ is a perceptual/feature-matching term (using a frozen encoder) to preserve anatomy/texture, $\mathcal{L}_{\text{ODE}}$ penalizes deviations from the latent ODE $\dot{Z}=f_{\theta}(Z,t)$ along the 24-month trajectory, and $\lambda_{\text{feat}},\lambda_{\text{ODE}}$ are scalar weights set to $0.1$ in our experiments. All models used the Adam optimizer (learning rate $10^{-4}$, weight decay $10^{-5}$, batch size 1 3D MRI) with batch normalization and early stopping on A100 GPUs. Each model was trained with a fixed maximum training time of 36 hours. Evaluation metrics were MSE, Peak Signal-to-Noise Ratio (PSNR), Structural Similarity Index Measure (SSIM). To study how well the models predict changes, we used $\Delta$-Pearson correlation for longitudinal change. Define predicted image by $\hat{x}^i_t$ and  vectorized in-mask change maps $\Delta^i=Vec(x^i_{t+24}-x^i_t)$ and $\hat{\Delta}^i=Vec(\hat{x}^i_{t+24}-x^i_t)$. Then participant-wise (Global) $\Delta$-Pearson correlation is the average Pearson correlation $(1/|Test|) \sum_{j \in Test} r(\hat{\Delta}^j,\Delta^j)$. 

To define \textit{Voxel-wise $\Delta$-Pearson correlation} for each voxel $v$, we compute $r\!\left(\{\widehat{\Delta}^{j}_v\}_j,\{\Delta^{j}_v\}_j\right)$ across participants and average the resulting $r$ over in-mask voxels.

\section{Results}
\label{sec:results}

\subsection{ADNI Test Set Performance}

Table \ref{tab:comprehensive_results} presents comprehensive results. On BigDataset, U$^2$-Net achieved best reconstruction (MSE=\num{3.27e-4}, PSNR=31.32 dB, SSIM=0.990), while ODE-UNet achieved highest change prediction ($\Delta$-Pearson=0.253). When trained on SmallDataset, UNETR achieved lowest MSE (\num{3.35e-4}), while ODE-UNet excelled at change prediction ($\Delta$-Pearson=0.215).

\begin{table*}[t]
\centering
\scriptsize
\setlength{\tabcolsep}{3pt}
\renewcommand{\arraystretch}{1.12}
\resizebox{\textwidth}{!}{%
\begin{tabular}{@{}l c c c c c c c c c c c@{}}
\hline
\multicolumn{2}{c}{} &
\multicolumn{5}{c}{\textbf{ADNI Test Set}} &
\multicolumn{5}{c}{\textbf{AIBL Test Set}} \\
\cline{3-12}
\textbf{Model} & \textbf{Train} &
\textbf{MSE} & \textbf{PSNR} & \textbf{SSIM} &
\textbf{\shortstack{$\Delta$-Pearson\\ (Global)}} &
\textbf{\shortstack{$\Delta$-Pearson\\ (Voxel)}} &
\textbf{MSE} & \textbf{PSNR} & \textbf{SSIM} &
\textbf{\shortstack{$\Delta$-Pearson\\ (Global)}} &
\textbf{\shortstack{$\Delta$-Pearson\\ (Voxel)}} \\
\hline
UNet        & Big   & 4.23 & $30.34\pm2.99$ & 0.987 & 0.202 & $0.166\pm0.091$ & 3.35 & $31.40\pm3.09$ & 0.991 & \textbf{0.209} & \textbf{$0.201\pm0.086$} \\
U$^2$-Net   & Big   & \textbf{3.27} & \textbf{$31.32\pm2.48$} & \textbf{0.990} & 0.241 & $0.185\pm0.131$ & \textbf{2.37} & \textbf{$33.25\pm2.64$} & \textbf{0.993} & 0.169 & $0.144\pm0.123$ \\
UNETR       & Big   & 3.39 & $30.98\pm2.46$ & 0.989 & 0.231 & $0.197\pm0.099$ & 2.47 & $32.27\pm2.62$ & 0.992 & 0.174 & $0.169\pm0.091$ \\
TEUNET      & Big   & 5.41 & $29.47\pm3.18$ & 0.988 & 0.126 & $0.161\pm0.102$ & 3.40 & $31.80\pm3.23$ & 0.992 & 0.180 & $0.219\pm0.092$ \\
ODE-UNet    & Big   & 5.02 & $29.09\pm2.51$ & 0.987 & \textbf{0.253} & \textbf{$0.227\pm0.111$} & 4.97 & $29.16\pm2.63$ & 0.989 & 0.186 & $0.191\pm0.104$ \\
\hline
UNet        & Small & 4.31 & $30.39\pm2.96$ & 0.987 & 0.100 & $0.093\pm0.092$ & 2.94 & $32.24\pm3.11$ & 0.991 & 0.170 & $0.160\pm0.084$ \\
U$^2$-Net   & Small & 3.44 & \textbf{$31.04\pm2.41$} & 0.989 & 0.171 & $0.167\pm0.117$ & 2.66 & $32.64\pm2.61$ & 0.993 & 0.133 & $0.123\pm0.110$ \\
UNETR       & Small & \textbf{3.35} & $31.01\pm2.47$ & 0.989 & 0.194 & $0.173\pm0.101$ & 2.49 & $32.31\pm2.65$ & 0.992 & 0.175 & $0.156\pm0.095$ \\
TEUNET      & Small & 5.26 & $29.04\pm2.86$ & 0.986 & 0.190 & $0.181\pm0.114$ & 4.98 & $29.04\pm2.47$ & 0.989 & 0.140 & $0.169\pm0.101$ \\
ODE-UNet    & Small & 3.53 & $31.01\pm2.60$ & \textbf{0.990} & \textbf{0.215} & \textbf{$0.226\pm0.107$} & \textbf{2.21} & \textbf{$33.33\pm2.73$} & \textbf{0.994} & \textbf{0.205} & \textbf{$0.205\pm0.095$} \\
\hline
\end{tabular}%
}
\vspace{0.3em}
\caption{Performance metrics on ADNI and AIBL test sets. \textbf{MSE values are reported as $\times 10^{-4}$} (i.e., actual MSE $=$ shown value $\times 10^{-4}$). Values for PNSR and voxel-wise $\Delta$-Pearson are reported as mean $\pm$ standard deviation across participants and in-mask voxels, respectively.}
\label{tab:comprehensive_results}
\end{table*}

\subsection{AIBL External Validation}

\begin{figure}[h]
\centering
\includegraphics[width=\columnwidth]{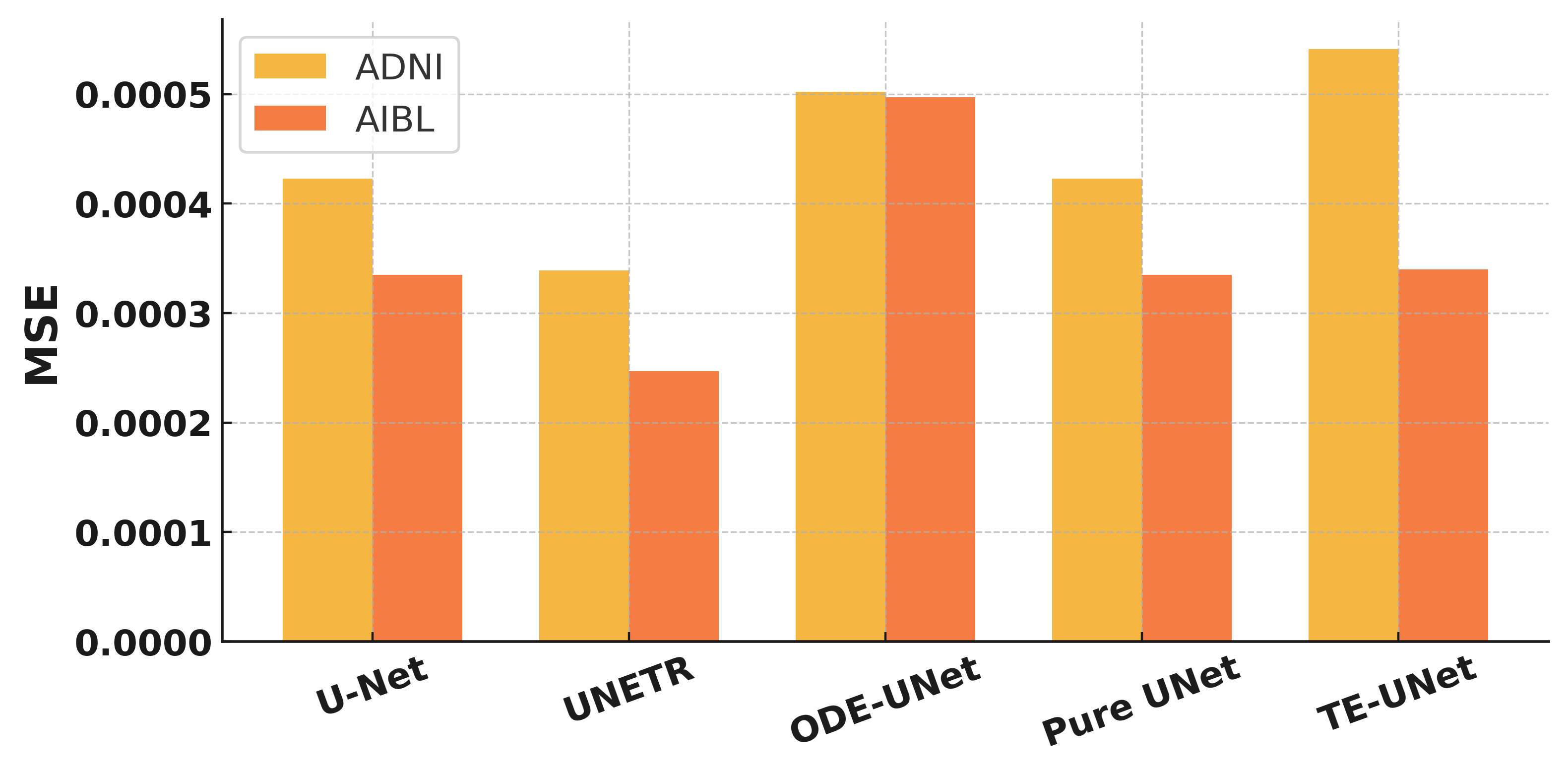}
\caption{MSE comparison between ADNI and AIBL test sets.}
\label{fig:adni_aibl}
\end{figure}

Models generalized robustly to AIBL (Table \ref{tab:comprehensive_results}). ODE-UNet trained on SmallDataset achieved exceptional results (MSE=\num{2.21e-4}, PSNR=33.33 dB, SSIM=0.994, $\Delta$-Pearson=0.205), surpassing ADNI performance. Figure \ref{fig:adni_aibl} illustrates consistent cross-dataset performance.

\subsection{Error Analysis}

\begin{figure*}[t]
\centering
\includegraphics[width=0.85\textwidth]{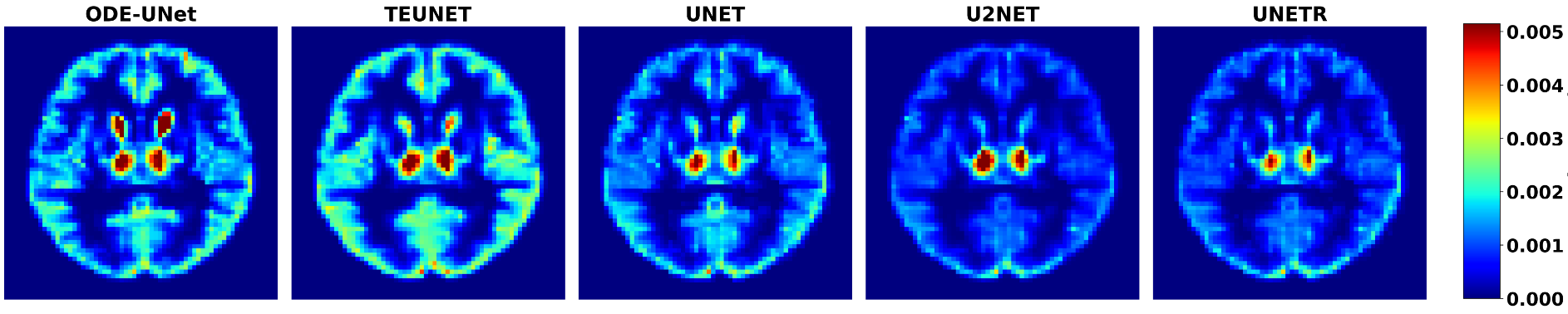}
\caption{Voxel-wise MSE maps averaged across ADNI test participants. Axial slices at $z = 14mm$  of the MNI space are shown.}
\label{fig:axial_mse}
\end{figure*}

\begin{figure}[h]
\centering
\includegraphics[width=\columnwidth]{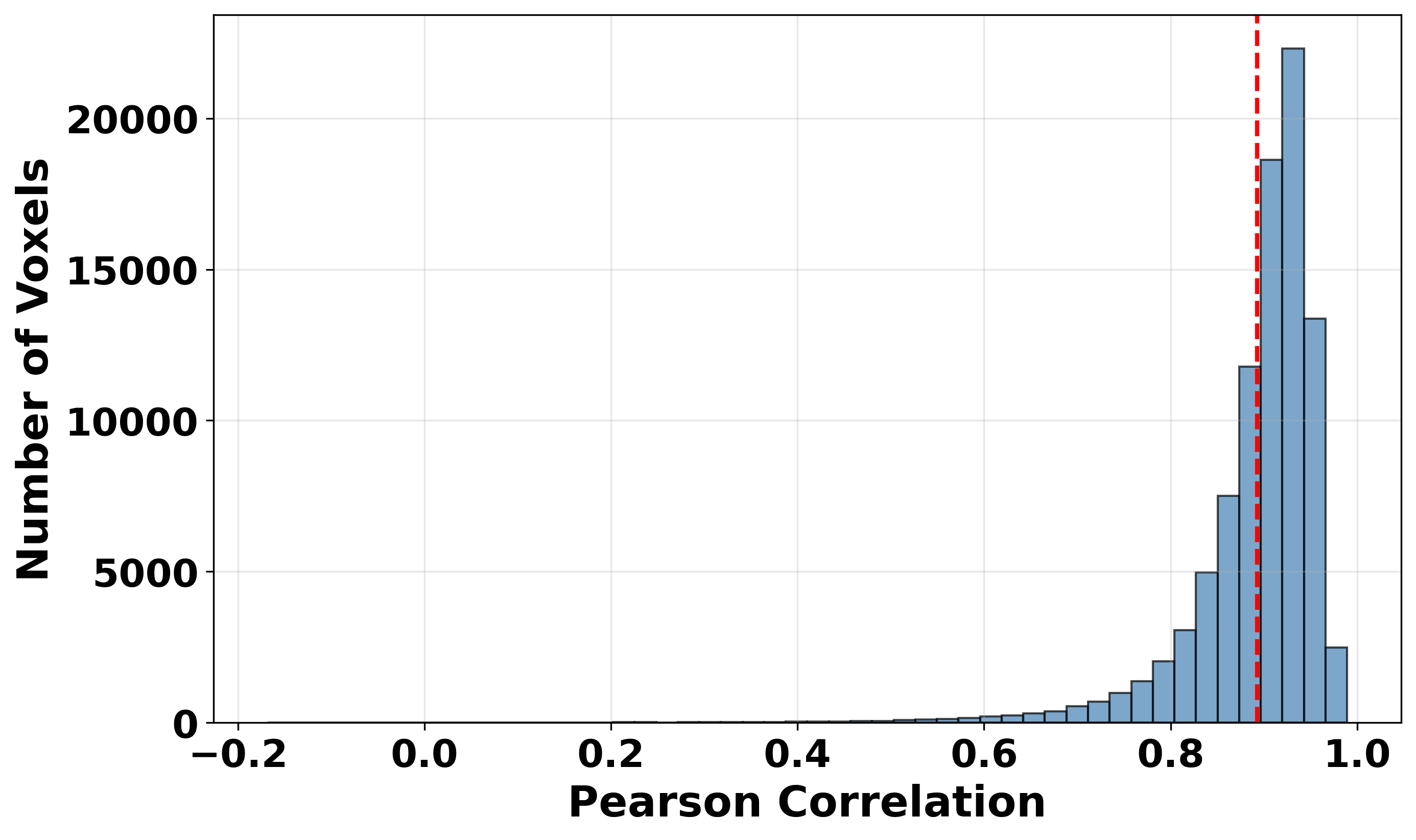}
\caption{Histogram of voxel-wise correlations between the predicted 24-month image and the actual 24-month image across participants in  ADNI test set (ODE-UNet trained on BigDataset).}
\label{fig:hist_intensity}
\end{figure}

\begin{figure}[h]
\centering
\includegraphics[width=\columnwidth]{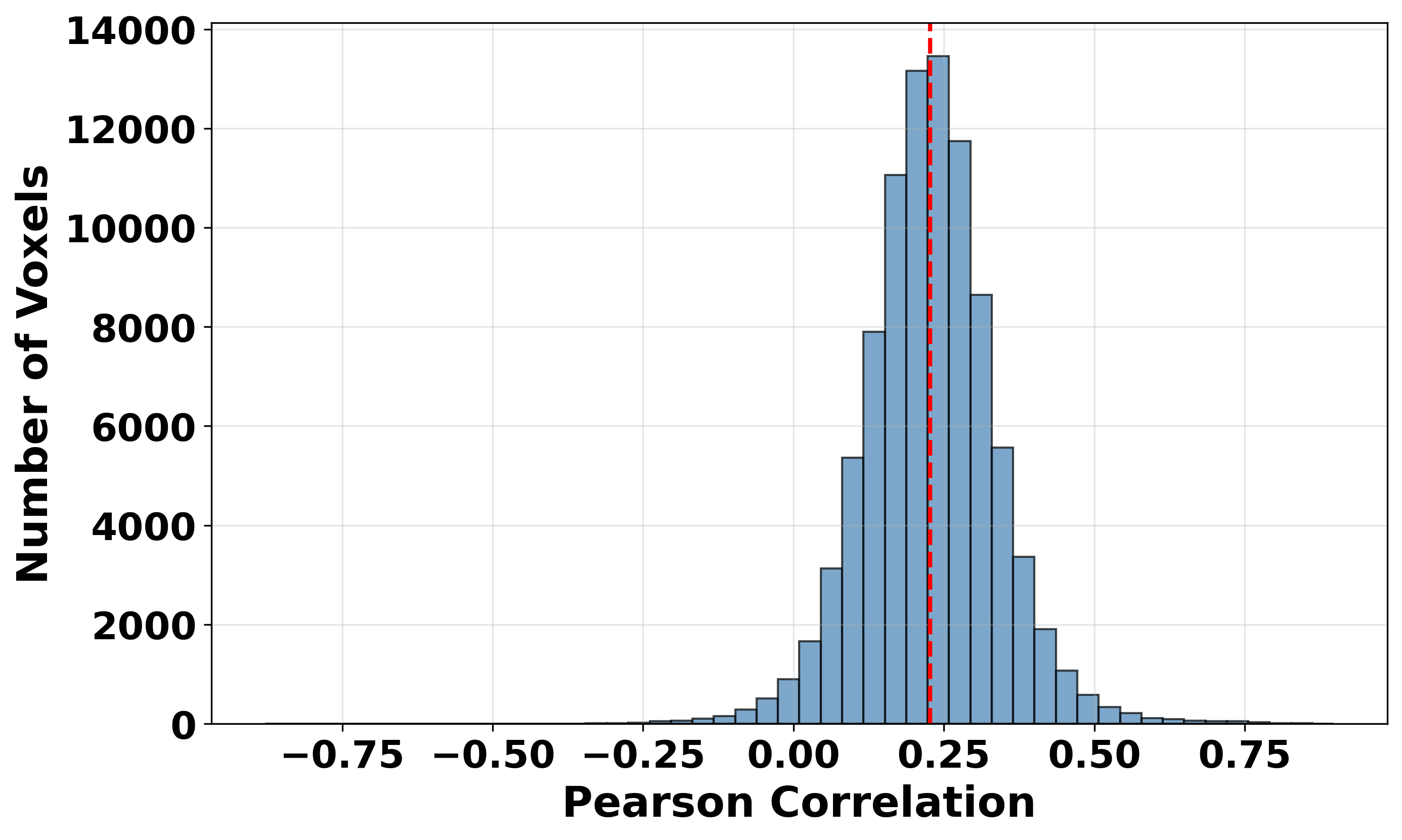}
\caption{Histogram of voxel-wise $\Delta$-Pearson correlations on ADNI test set (ODE-UNet trained on BigDataset).}
\label{fig:hist_delta}
\end{figure}

Figure \ref{fig:axial_mse} shows U$^2$-Net exhibited most concentrated errors, with all models showing higher errors in rapid-change regions. Figures \ref{fig:hist_intensity} and \ref{fig:hist_delta} demonstrate ODE-UNet's preservation of intensity distributions and capture of longitudinal change patterns.

\section{Discussion}
\label{sec:conclusion}

This work presents a comprehensive comparison among five deep learning architectures for longitudinal, voxel-level MRI prediction across participants with varying levels of cognitive impairment (none to dementia). Given a single baseline, models predict future MRI scans that capture spatially complex patterns of potential neurodegeneration. U\textsuperscript{2}\!-Net provided the best overall image fidelity in three of the four Train\,$\times$\,Dataset blocks (highest PSNR and lowest/near-lowest MSE), while ODE\mbox{-}UNet was competitive—being best on AIBL/Small (PSNR \(33.33\pm2.73\) dB, SSIM \(0.994\)). UNETR achieved the lowest MSE only on ADNI/Small. On the other hand, ODE-UNet performed best on change prediction, with \(\Delta\)-correlation values as high as 0.253, underlining the strength of continuous-time modeling with respect to disease evolution.
Taken together, these results indicate that MSE alone may not be a reliable indicator of predictive quality: models can obtain lower MSE yet produce less realistic or less temporally consistent longitudinal predictions. For example, although U\textsuperscript{2}-Net and UNETR sometimes achieved lower MSE, ODE-UNet yielded higher \(\Delta\)-Pearson correlation (global and voxel-wise), indicating better agreement with longitudinal change.

On average, under a fixed 36\,h training budget, models trained on the more diverse BigDataset performed best especially for change prediction, where ODE\mbox{-}UNet achieved higher $\Delta$-Pearson scores both globally and at the voxel level. In contrast, SmallDataset training sometimes resulted in lower MSE, largely because its smaller size allowed models to fully converge within the time limit, whereas BigDataset runs remained under-optimized for some models. Thus, the observed MSE difference primarily reflects the constrained optimization time rather than any inherent limitation of BigDataset. All models generalized very well to the independent AIBL cohort, indicating that learned patterns reflect core aspects of brain aging. Among them, U\textsuperscript{2}-Net and ODE-UNet demonstrated the smallest errors on AIBL, thus supporting robustness across populations.

Voxel-level prediction allows advantages over regional biomarker modeling, such as preserving spatial dependencies and enabling individualized downstream analyses. Corresponding error maps and histograms confirmed low reconstruction error and effective modeling of both subtle and gross changes.

We made certain methodological choices that warrant commenting. We chose to use spatially normalized and modulated gray matter density (GMD) maps rather than raw MRIs as GMD can be considered to be quantitative in the sense that GMDs from two individuals can be compared. We trained models on a fixed 24-month prediction for both methodological and conceptual simplicity. We chose to downsample images for computational reasons. Prediction of raw images, longer and variable horizons, multi-modal inputs, e.g. integration of demographic and genetic information, and uncertainty estimation are left for future work. Our results suggest that deep learning has the potential to provide individualized forecasting of brain atrophy. Provided that such models are further refined and validated, they will have significant implications for prognosis, improvement in clinical trial design, and personalized planning of interventions for neurodegenerative disease.

\vspace{0.5\baselineskip}
{\centering\bfseries Compliance with Ethical Standards\par}
\noindent This retrospective study used data from ADNI (\url{http://adni.loni.usc.edu}) and AIBL (\url{https://aibl.org.au}), which received ethical approval with participant consent. Secondary analysis required no additional approval per data use agreements.

\vspace{0.5\baselineskip}
{\centering\bfseries Acknowledgments \par}
\noindent 
Data used in the preparation of this article were obtained from the Alzheimer's Disease Neuroimaging Initiative (ADNI) database (\url{http://adni.loni.usc.edu}) and the Australian Imaging Biomarkers and Lifestyle flagship study of ageing (AIBL) funded by the Commonwealth Scientific and Industrial Research Organisation (CSIRO), which was made available at the ADNI database. The investigators within ADNI and AIBL contributed to the design and implementation of ADNI/AIBL and/or provided data but did not participate in analysis or writing of this report. A complete listing of ADNI investigators can be found at \url{http://adni.loni.usc.edu/wp-content/uploads/how_to_apply/ADNI_Acknowledgement_List.pdf}. AIBL researchers are listed at \url{http://www.aibl.csiro.au}

Data collection and sharing for ADNI were supported by the Alzheimer's Disease Neuroimaging Initiative (ADNI) (National Institutes of Health Grant U01 AG024904) and DOD ADNI (Department of Defense award number W81XWH-12-2-0012). ADNI is funded
by the National Institute on Aging, the National Institute of Biomedical Imaging and
Bioengineering, and through generous contributions from the following: AbbVie, Alzheimer’s Association; Alzheimer’s Drug Discovery Foundation; Araclon Biotech; BioClinica, Inc.; Biogen; Bristol-Myers Squibb Company; CereSpir, Inc.; Cogstate; Eisai Inc.; Elan Pharmaceuticals, Inc.; Eli Lilly and Company; EuroImmun; F. Hoffmann-La Roche Ltd and its affiliated company Genentech, Inc.; Fujirebio; GE Healthcare; IXICO Ltd.; Janssen Alzheimer Immunotherapy Research \& Development, LLC.; Johnson \& Johnson Pharmaceutical Research \& Development LLC.; Lumosity; Lundbeck; Merck \& Co., Inc.; Meso Scale Diagnostics, LLC.; NeuroRx Research; Neurotrack Technologies; Novartis Pharmaceuticals Corporation; Pfizer Inc.; Piramal Imaging; Servier; Takeda Pharmaceutical
Company; and Transition Therapeutics. The Canadian Institutes of Health Research is
providing funds to support ADNI clinical sites in Canada. Private sector contributions are facilitated by the Foundation for the National Institutes of Health ( www.fnih.org). The grantee organization is the Northern California Institute for Research and Education, and the study is coordinated by the Alzheimer’s Therapeutic Research Institute at the University of Southern California. ADNI data are disseminated by the Laboratory for Neuro Imaging at the University of Southern California.

Supported by The Research Council of Finland grant 351849 (under the frame of ERA PerMed ("Pattern-Cog"), 346934 (PRIMAL), and  358944 (Flagship of Advanced Mathematics for Sensing Imaging and Modeling) and by the Finnish Ministry of Education and Culture’s Pilot for Doctoral Programmes (Pilot project Mathematics of Sensing, Imaging and Modelling). We acknowledge CSC – IT Center for Science, Finland and UEF Bioinformatics Center. 

We used OpenAI’s GPT-5 for language editing and general readability during manuscript preparation. The authors reviewed and revised the text subsequently, so they are solely responsible for the final content.

\bibliographystyle{IEEEbib}
\bibliography{refs}

\end{document}